# Exact Blur Measure Outperforms Conventional Learned Features for Depth Finding

Akbar Saadat
Passive Defence R&D Dept.
Tech. Deputy of Iranian Railways
Tehran, Iran

*Abstract*—Image analysis methods that are based on exact blur values are faced with the computational complexities due to blur measurement error. This atmosphere encourages scholars to look for handcrafted and learned features for finding depth from a single image. This paper introduces a novel exact realization for blur measures on digital images and implements it on a new measure of defocus Gaussian blur at edge points in Depth From Defocus (DFD) methods with the potential to change this atmosphere. The experiments on real images indicate superiority of the proposed measure in error performance over conventional learned features in the state-of the-art single image based depth estimation methods.

## I. Introduction

Blur measurement error increases the computational complexity of DFD, and encourages the community of research to look for intuitive nondeterministic approaches in artificial intelligence to find depth of a scene from its images. The most related field under this demand is depth oriented image segmentation. In contrast to DFD this approach does not consider the relative blur between two images due to depth, but invests on the labelling the depth in a single image. This approach has involved with high complexities for depth finding.

For inferring depth from a single image a complete database of the world images is required with their 3-D coordinates. This is running in literature and highlighted in [1] for integration all present RGB Depth databases. The research in [2] exploited the availability of a pool of images with known depth to formulate depth estimation as an optimization problem. The methods for estimating depth from a single image are commissioned to touch the human skills on inferring 3D structure or depth. The evolution road of the present methods, which is tightened by enforcing geometric assumptions to infer the spatial layout of a room in [3] and [4] or outdoor scenes in [5], is being expanded by handcrafted features in [2], [6], [7], [8] and [9] for more general scenes. Most limitations are supposed to be diminished by learning features in the multi layers of Convolutional Neural Networks (CNN) [10], [11], [12] which infer directly depth map from the image pixel values. In the current literature, there is no basic difference between depth estimation and semantic labelling, as jointly performing both can benefit each other [7]. The possibility of generating semantic labelling with the information to guide depth perception given in [13] supports the legality of the multistage inferring process in CNN.

Both handcrafted and learned features based depth estimation from a single image methods, suffer from high complexity in depth inferring process and in adapting (tuning or training) the model parameters. Changing the model from handcrafted features to CNN shifts the complexity in the depth inferring process from retrieval time to calculation architecture. Both the adaptation and the inferring complexities increase with the number and size of the input images. However, DFD methods are able to provide a closed form formula for depth of a local image independent of the size and number of the input images, with insignificant processing time. These advantages can switch back depth finding methods from handcrafted and learned features to DFD measures, provided that both approaches lead to comparable results. This statement is quantified here in terms of measurement error by improving the blur measures in DFD.

Regardless of all environmental sources of error in DFD, formulating blur measurement is expected to be free of internal error. Inductive replacing differentials with differences on discrete implementations is conventional source of the blur measurement error in DFD formulation. This error should be eliminated for effective comparing the performance of DFD against conventional learned features in single image based methods

This paper contributes the DFD methods first by introducing the exact discrete realization of a general blur measure on digital images, and then by presenting a new blur measure in the exact form with interesting results for comparing DFD with the single image based methods. Problem formulation is based on the image formation model given in the next section. In the following sections the exact discrete realization is applied on a well-accepted conventional blur measure in literature and the improvement due to that is quantified. Then, the new blur measure is induced of the exact realisation for the present blur measure. The proposed blur measure is compared first with the exact realisation and then with the state-of the-art single image based depth estimation methods over the test images of the Make3D range image dataset [8]. Comparison results simplify the decision for selecting the proposed measure of blur or conventional learned features for depth finding.

## II. DFD Image Formation Model

DFD obtains depth by modelling the depth dependent blur or defocus Point Spread Function (PSF). The method obtains depth by estimating the scale of the PSF at each image point using a raw blur measure. In DFD theory, the defocused image of a scene point $i(x,y)$ is obtained by convolving the focused image $s(x,y)$ with the PSF $h(x,y)$ as

$$i(x,y) = s(x,y) * h(x,y), \quad (1)$$

The blur parameter $\sigma$ is a space-variant that represents depth variations over the scene. Relying on the central limit theorem, it is usually assumed that the defocus PSF is a Gaussian function as

$$h(x,y) = \frac{1}{2\pi\,\sigma^2(x,y)} e^{-\frac{x^2+y^2}{2\sigma^2(x,y)}}. \quad (2)$$

The analytic approaches in DFD obtain depth by solving the equations of the blur values over two images of a scene at different settings for the imaging system. In most general cases there are two equations: the first is a linear equation that depends on the camera settings[14], and the second one sets the difference between the squared blur values to its analytic measures over the images. These are called camera-based and image-based DFD equation pairs, with the solutions for the objective blur or depth dependent blur of both image points. The image based equation is obtained by local computing on the images based on an analysis in the frequency( [15] , [16] and [17]) or spatial domain ([[18], [19] and [20]).

In the exceptional cases of the scene focussed image with the step edges [21], with sharp textures [22], or with the gradient described by the white Gaussian distribution random process [23], one image will be enough to measure the objective blur, the depth dependent blur, or the likelihood of a candidate defocus scale, respectively. Modelling the image gradient by the white Gaussian random process is a creative technique for the blur estimation in a closed form, regard to all dependencies to the image contents. Although, the method is enriched by smoothness and colour edge information, its applications does not extends beyond the labelling for foreground/background segmentation [24]. This method does not measure, but selects the maximum likelihood local defocus scale of a given set, and it does not guarantee labelling the image patches by true values of scale.

In [21], the blur measure at the edge locations is related analytically to the gradient between the input image and re-blurred version of that. These images are replaced with two different re-blurred versions of the input image in [25] and [26] by hard assumption on the PSF which is confirmed more by the smaller values of blur. It will be shown that this range of blur values are faced with the most measurement errors.

This paper locates a point in an image by Canny edge detector and validates it for measuring the blur when there is just one edge orientation inside the respected measurement circle which is considered with the radius of three pixel width centred on that point. A valid local image for measuring the blur is modelled by a step edge that is defocused consecutively by two Gaussian blur functions. First with the Inherent or subjective blur $\sigma_s$ that makes the original focused image of a scene with non-sharp edge, then with the depth dependent or objective blur $\sigma_o$. The result is similar to blurring the step edge with the Gaussian PSF by the absolute blur $\sigma = \sqrt{\sigma_s^2 + \sigma_o^2}$. In this model the operator of the DFD image based equation makes equal outputs for the absolute and for the depth dependent blur values over two given images of a local area with the same $\sigma_s$. Therefore, an estimator of the absolute blur value is enough for the image based DFD equation.

To measure the absolute blur value over an image, a circle with the radius of three pixel width for the local measurements is centred at each pixel, and it is assumed that the blur value is fixed over that. The blur measure operator is applied to all the validated local images. Including the subjective blur in the objective, the focused image with the grey level range $(i_{min}, i_{max})$ of a valid local image can be described by the 2-D step function as $s(x,y) = (i_{max}-i_{min})U(y) + i_{min}$ in the local image. Convolving this by the defocus PSF with $\sigma(x,y) = \sigma$ in (2), makes the defocused image given by $i(y)$ as

$$\begin{aligned}i(x,y) &= s(x,y) * h(x,y) \\ &= i_{min} + \frac{i_{max}-i_{min}}{2}\left(1 + erf\left(\frac{y}{\sqrt{2}\sigma}\right)\right) \triangleq i(y),\end{aligned} \quad (3)$$

where $erf$ is the error function defined by (4).

$$erf(x) = \frac{1}{\sqrt{\pi}}\int_{-x}^{x} e^{-t^2} dt \quad (4)$$

## III. Conventional and Exact Blur Measures

This section abstracts the analytic blur measure in [21] to introduce the exact discrete value for that, then modifies the result to a new blur measure with less complexity and higher error performance. The proposed measure in [21] is based on the gradient ratio between the input image and re-blurred version of that. The magnitude of the gradient of $i(y)$ in (3) is

$$|\nabla i(y)| = \frac{\partial i(y)}{\partial y} = \frac{i_{max}-i_{min}}{\sqrt{2\pi}\sigma}\exp\left(\frac{-y^2}{2\sigma^2}\right) \quad (5)$$

and the re-blurred version of $i(y)$ by the Gaussian kernel with the standard deviation $\sigma_1$ is (6).

$$i_1(y) = i_{min} + \frac{i_{max}-i_{min}}{2}\left(1 + erf\left(\frac{y}{\sqrt{2\pi(\sigma^2+\sigma_1^2)}}\right)\right) \quad (6)$$

The magnitude of the gradient of $i_1(y)$ is obtained as (7)

$$|\nabla i_1(y)| = \frac{i_{max}-i_{min}}{\sqrt{2\pi(\sigma^2+\sigma_1^2)}}\exp\left(\frac{-y^2}{2(\sigma^2+\sigma_1^2)}\right) \quad (7)$$

The gradient ratio between the input and re-blurred images at the edge location $y = 0$ leads to (8).

$$R_G(\sigma) = \frac{|\nabla i(0)|}{|\nabla i_1(0)|} = \sqrt{\frac{\sigma^2 + \sigma_1^2}{\sigma^2}} \quad (8)$$

With the known value of $\sigma_1$ the blur value is estimated by (9).

$$\sigma_{RG}(R_G) = \frac{\sigma_1}{\sqrt{R_G^2 - 1}} \quad (9)$$

The exact discrete value of $R_G(\sigma)$ is obtained as $R_{Gd}(\sigma)$ in (10), by using the discrete approximation of the absolute gradient in (8).

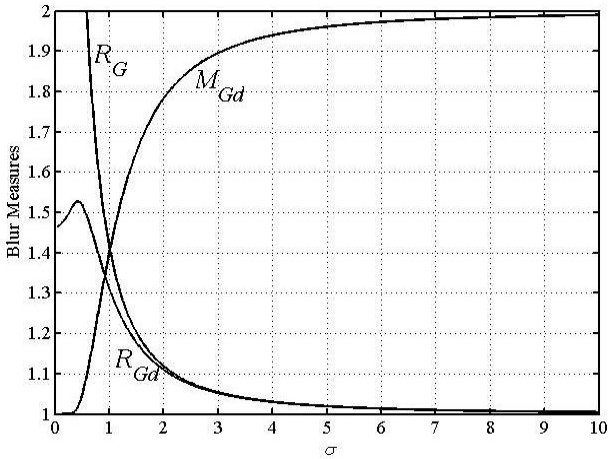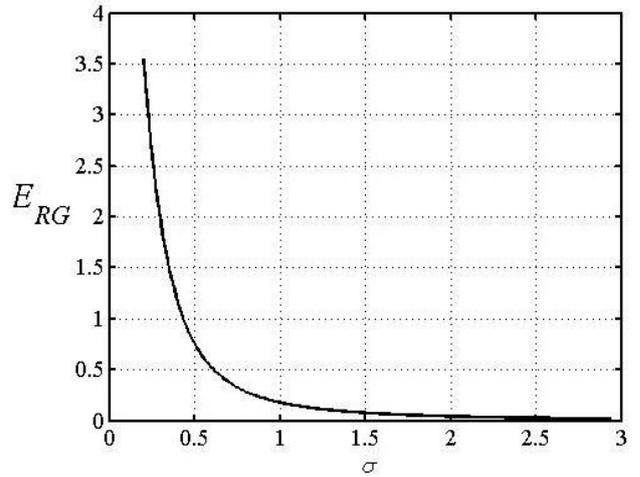

Fig. 1. (a) Variation of blur measures versus the blur values. (b) The blur measurement error caused by conventional discretizing the blur measure $R_G$.

$$R_{Gd}(\sigma) = \frac{i(1)-i(0)}{i_1(1)-i_1(0)} = \frac{erf\left(\frac{1}{\sqrt{2}\sigma}\right)}{erf\left(\frac{1}{\sqrt{2(\sigma^2+\sigma_1^2)}}\right)} \quad (10)$$

The blur value is estimated by applying the inverse function of $R_{Gd}(\sigma)$ on the computed $R_{Gd}$ as (11).

$$\sigma_{RGd} = R_{Gd}^{-1}(R_{Gd}) \quad (11)$$

For the noise free image with the proposed model of image formation the relation $R_{Gd} = R_{Gd}(\sigma)$ leads to $\sigma_{RGd} = \sigma$. In the case of noisy images $\sigma_{RGd}$ will be deviated from $\sigma$ by noise effects.

This paper introduces a new blur measure similar to $R_G$ without re-blurring input image. The concept of variation of the re-blurred image in the asymmetric range (0,1) in the denominator of the fraction for $R_{Gd}$ in (10) can be replaced with that of the input image in the symmetric range (−1,1). For the corresponding nominator which is needed to be greater than the denominator, the variation range (−2,2) is a good candidate. Following this induction, the new measure of blur with the exact discrete format is introduced as (12).

$$M_{Gd}(\sigma) = \frac{i(2)-i(-2)}{i(1)-i(-1)} = \frac{erf\left(\frac{\sqrt{2}}{\sigma}\right)}{erf\left(\frac{1}{\sqrt{2}\sigma}\right)} \quad (12)$$

The blur value is estimated by the reverse function in (13).

$$\sigma_{MGd} = M_{Gd}^{-1}(M_{Gd}) \quad (13)$$

Implementing (12) requires edge locating which is done by Canny edge detector, and edge orientation which is done by maximising the image intensity variations inside the measurement circle.

The blur measure $R_G(\sigma)$, its exact discrete value $R_{Gd}(\sigma)$ and the new blur measure $M_{Gd}(\sigma)$ are plotted in Fig. 1.a. $R_{Gd}(\sigma)$ and $M_{Gd}(\sigma)$ in (11) and (13) make the exact value of σ from monotonic functions of σ for σ > 0.42 pixel width. More precisely for the exact values of blur measures, $R_{Gd}$ for σ > 0.42 and $M_{Gd}$ for σ > 0 are monotonic. In the monotone range all blur measures are invertible, but just the inverse of $R_{Gd}(\sigma)$ and $M_{Gd}(\sigma)$ will make the exact value of σ from discrete local image samples. The results in Fig. 1.a indicate unbounded difference between the blur measure $R_G$ and its exact discrete value $R_{Gd}$ for infinite small values of σ. Therefore, the mistake on using $R_G$ instead of $R_{Gd}$ will lead to catastrophic measurement error on estimating the small values of σ from discrete local image samples.

The blur measures $R_{Gd}(\sigma)$ and $M_{Gd}(\sigma)$ can be compared graphically in Fig. 1.a. The size of both the variation range, and the monotone range enclosed by the variation range indicate the power of resolving blur values by the measures. For the both cases the proposed blur measure $M_{Gd}$ advances on $R_{Gd}$. The size of variation range of $M_{Gd}(\sigma)$ is approximately twice that of $R_{Gd}(\sigma)$. This feature and no need to re blurring the input image indicate the preference of $M_{Gd}(\sigma)$ to $R_{Gd}(\sigma)$ for depth finding.

IV. PROMOTION IN ERROR PERFORMANCE

Replacing any conventional blur measure with its exact discrete value in blur estimation reduces the blur measurement error. The amount of error reduction is quantified for the blur measure $R_G$ as follows. The error is caused by applying (9) for estimating σ with the discrete blur measure $R_{Gd}$. Therefore, the relative blur measurement will be given by (14),

$$E_{RG}(\sigma) = \left|\frac{\sigma_{RG}(R_{Gd})-\sigma}{\sigma}\right| = \left|\frac{\sigma_1/\sigma}{\sqrt{R_{Gd}^2(\sigma)-1}} - 1\right| \quad (14)$$

As $R_{Gd}$ in (10) is a known function of the blur value σ, in the proposed image formation model $E_{RG}$ will be a known function of σ and the reblurring parameter $\sigma_1$.

For other image formation models and in the presence of noise, $R_{Gd}$ in (10) is a known function of the image samples and $E_{RG}$ as a function of $R_{Gd}$ in (14) represents the measurement error caused by the image noise and deviation from the proposed model. In this case $R_G$ could be assumed as an heuristic blur measure free of Gaussian assumption on the PSF. $E_{RG}$, as shown in Fig. 1.b, is unbounded in accordance with $R_G$ for infinite small values of blur, and is not fairly low in a wide range of blur values. Therefore its average on any blur range expressed by $(0, \sigma_{max})$ is not theoretically limited.

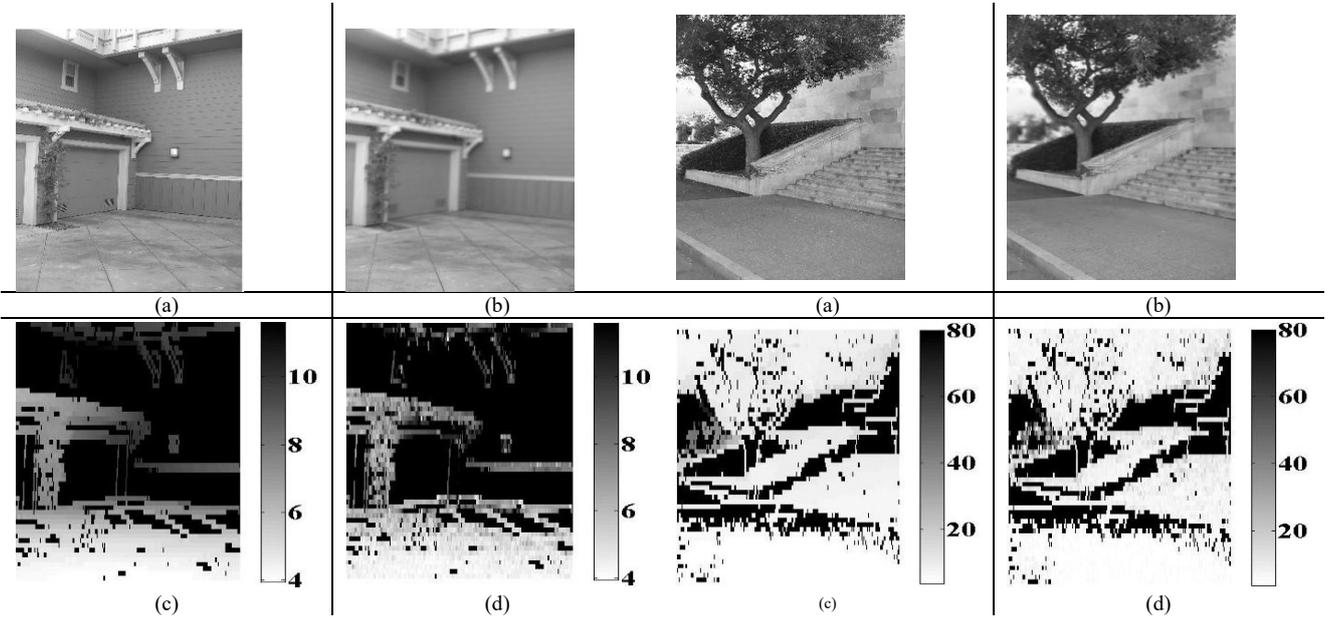

Fig. 2. Experiment results on two sample of the test image of the Make3D range image dataset. (a) Original image in size of 2272 by1704 pixels. (b) The defocused image by depth dependent blur. (c) Original depth map in size of 55 by 305 pixels. (d) Estimated depth map by the proposed measure in the size of original depth map.

## V. EXPERIMENT RESULTS

The analytic results in the previous sections demonstrated theoretically superiority of the exact discrete values of blur measures over the conventional ones and that of the proposed measure over the present one for the images described by (3) in absence of noise. An experiment is planned on real images to compare the exact discrete value of the proposed blur measure with the state-of the-art single image based depth estimation methods. More specifically, the error performance of $M_{Gd}$ over the test images of the Make3D range image dataset [8] is obtained to compare with the results reported by these methods. The dataset contains images of natural scenes in the size of 2272x1704 and corresponding depth maps with the resolution of 55x305. Therefore, depth values are assumed to be known in none overlapping rectangular 41x5 superpixels that cover the rectangle area in size of 2256x1526 in the middle part of the given images. This dataset has been experimented in literature and the mean absolute relative error has been reported 53% in [27], 37% in [8], 37.5% in [2], 36.2% in [6], 33.8% in [9], and 30.7% in [10] and [11], for that.

Given an RGBD input image for the blur measurement, the depth map of that denoted by $D(x,y)$ is applied for making a space-variant blur parameter $\sigma(x,y)$ by

$$\sigma(x,y) = c + d\, log(D(x,y)). \quad (15)$$

This model provides depth independent error for $\sigma$ as $\Delta\sigma$ due to the error in the range finding for depth map when the relative error $\Delta D/D$ is fixed over the whole range of $D$. The model can be fitted locally on the model $\hat{c} + \hat{d}/D(x,y)$ in Geometric Optics for applying with the camera settings. The parameters c and d are determined by aligning the extreme values of $\sigma$ and D in the ranges $\sigma_{min} \leq \sigma \leq \sigma_{max}$ and $D_{min} \leq D \leq D_{max}$ by the corresponding pairs $(D_{min}, \sigma_{min})$ and $(D_{max}, \sigma_{max})$. $D_{min}$ and $D_{max}$ are available in the depth data, and $(\sigma_{min}, \sigma_{max})$ is set at $(0.5, 10)$ to ensure of the blur measure being in a proper monotone measurement range as in Fig. 1.a. The minimum value is in favour of less error in low blur values with the cost of less resolution for large values. The Gaussian blur function $h(x,y)$ in (2) is applied on the input image to obtain the defocused pair of the input image and make the DFD image pairs. Based on two measures of blur, $M_{Gd1}$ over the original input image and $M_{Gd2}$ over the defocused image, the objective blur σ of the defocused image due to depth at the edge points is estimated by

$$\hat{\sigma}^2(x,y) = \sigma_{MGd}{}^2\big(M_{Gd2}(x,y)\big) - \sigma_{MGd}{}^2\big(M_{Gd1}(x,y)\big). \quad (16)$$

And, the estimation of the depth map at the edge points of the original input image will be

$$\widehat{D}(x,y) = \mathrm{Exp}\left(\frac{\hat{\sigma}(x,y)-c}{d}\right). \quad (17)$$

For obtaining the depth map with same resolution as the original one, the depth values over all pixels of each superpixel is set to the average of the estimated values at edge locations, or set to $D_{max}$ for those superpixels without detected edge point.

The experiment was run over the whole test images of the Make3D dataset. The difference between the pixel resolution in images and the superpixel resolution in depth maps helps to increase the density of depth estimation at edge points. While the fraction of the valid points on the original RGB images was obtained in average less than 6%, the figure for the the depth maps was more than 57%. The mean absolute relative error over the depth map of all 134 test images was found as 27.5% in average. The result is better than that of the published experimental results citec before for the depth estimation methods from a single image.

Fig. 2 has grouped the experiment results on two samples of the test images in the half left and half right part of the figure. The original image of the dataset, the defocussed image by depth dependent blur, the original depth map, and the estimated depth map are shown in each group. For the edgeless superpixels the estimated depth values are set to $D_{max}$ in Fig. 2.d to attain convenient visualization on the results. The reduction in the number of valid points for depth finding is firmly compensated by propagating the detected edge points in each superpixel to the all pixels inside that with same depth value. For the experiment in the left part 3.1% of the pixels in the original image are detected as the edge points, while the depth map is estimated for more than 43.4% of the superpixels at the resolution of the original dept map. The mean absolute relative error is obtained as 11% over the mentioned superpixels. The corresponding figures for the right part is 6.4%, 71% and 18.8%.

## VI. Conclusion

Exact discrete formulation led to introduce the exact realisation for the present blur measure $R_G(\sigma)$ as $R_{Gd}(\sigma)$ and the new blur measure $M_{Gd}(\sigma)$ in the exact form. The amount of reduction in measurement error caused by the exact value of the bur measure $R_G(\sigma)$ was quantified in Fig. 1.b. Then, the proposed blur measure $M_{Gd}(\sigma)$ was compared in error performance with $R_{Gd}(\sigma)$ and with the conventional learned features in the state-of the-art single image based depth estimation methods. Experiment results demonstrated the superiority of the proposed measure over the exact value of the present measure and over the conventional learned features in literature.

## References

[1] B. C. Russell and A. Torralba, "Building a database of 3d scenes from user annotations," in Proc. IEEE Conf. Comp. Vis. Patt. Recogn., 2009.

[2] B. Liu, S. Gould, and D. Koller, "Single image depth estimation from predicted semantic labels," in Proc. IEEE Conf. Comp. Vis. Patt. Recogn., 2010.

[3] V. Hedau, D. Hoiem, and D. A. Forsyth, "Thinking inside the box: Using appearance models and context based on room geometry," in Proc. Eur. Conf. Comp. Vis., 2010.

[4] D. C. Lee, A. Gupta, M. Hebert, and T. Kanade, "Estimating spatial layout of rooms using volumetric reasoning about objects and surfaces," in Proc. Adv. Neural Inf. Process. Syst., 2010.

[5] A. Gupta, A. A. Efros, and M. Hebert, "Blocks world re visited: Image understanding using qualitative geometry and mechanics," in Proc. Eur. Conf. Comp. Vis., 2010.

[6] K. Karsch, C. Liu and S. B. Kang, "Depth Transfer: Depth Extraction from Video Using Non-Parametric Sampling," in IEEE Transactions on Pattern Analysis and Machine Intelligence, vol. 36, no. 11, pp. 2144-2158, Nov. 2014.

[7] L. Ladick, J. Shi, and M. Pollefeys, "Pulling things out of perspective," in Proc. IEEE Conf. Comp. Vis. Patt. Recogn.,2014.

[8] A. Saxena, M. Sun, and A. Y. Ng, " Make3D: Learning 3-D Scene Structure from a Single Still Image, "In IEEE Transactions on Pattern Analysis and Machine Intelligence PAMI, Volume: 31, Issue: 5, pp. 824 - 840, 2009.

[9] M. Liu, M. Salzmann, and X. He, "Discrete-continuous depth estimation from a single image," in Proc. IEEE Conf.Comp. Vis. Patt. Recogn., 2014.

[10] F. Liu, C. Shen, G. Lin, and I. Reid, "Learning Depth from Single Monocular Images Using Deep Convolutional Neural Fields," IEEE Transactions on Pattern Analysis and Machine Intelligence ,Volume: 38, Issue: 10, Oct. 1 2016.

[11] F. Liu, C. Shen and G. Lin, "Deep Convolutional Neural Fields for Depth Estimation From a Single image," Proc. IEEE Conf. on Computer Vision and Pattern Recognition CVPR, 2015.

[12] D. Eigen, C. Puhrsch, and R. Fergus, " Depth map prediction from a single image using a multi-scale deep network," from a single image using a multi-scale deep network," in Proc. Adv. Neural Inf. Process. Syst., 2014.

[13] C. Liu, J. Yuen, and A. Torralba, "Nonparametric Scene Parsing: Label Transfer via Dense Scene Alignment", Proc. IEEE Conf. on Computer Vision and Pattern Recognition, CVPR, 2009.

[14] A.N. Rajagopalan, S. Chaudhuri and M.Uma, "Depth Estimation and Image Restoration Using Defocused Stereo Pairs, " IEEE Trans. Pattern Anall. Machine Intell., vol.26, no.11, pp.1521-1525,Nov.2004. 2

[15] A.P.Pentland, "A New Sense for Depth of Field," IEEE Trans. Pattern. Anal. Machine Intell., vol.PAMI-9, no.4, pp.523-531, July. 1987.

[16] M. Watanabe and S.K. Nayar, "Rational filters for passive depth from defocus, " International Journal of Computer Vision, 27 (3) (1998) 203-225.

[17] A. N. J. Raj and R.C. Staunton, " Rational Filters Design for Depth from Defocus, " Pattern Recognition, Vol. 45, No. 1, pp. 198-207, Jan. 2012

[18] J.Ens and P.Lawerence, ``An Investigation of Methods for Determining depth from Focus,'' IEEE Trans. Pattern Anal. Machine Intell., vol.15, no.2, pp.97-108, Feb.1993.

[19] M.Subbarao and G.Surya , "Depth From Defocus: A Spatial Domain Approach, " Int. Jour. Comput. Vision, vol.13, no.3, pp.271-294, 1994. 3

[20] T. Xian and M. Subbarao. Depth-from-defocus: Blur equalization technique. SPIE, 6382, 2006. 2

[21] S. Zhuo and T. Sim, " Defocus Map Estimation From a Single Image, " Pattern Recognition Volume 44 Issue 9, pp. 1852-1858, September, 2011.

[22] J. Lin, X. Ji, W. Xu, and Q. Dai, "Absolute Depth Estimation from a Single Defocused Image," IEEE Trans. Image Processing, vol. 22. Issue 11, pp. 4545-4550,Nov. 2013.

[23] A. Chakrabarti, T. Zickler, and W. T. Freeman, "Analyzing spatially varying blur," in Proc. IEEE CVPR, pp. 2512–2519, Jun. 2010.

[24] Zhu Xiang, S. Cohen, S. Schiller, and P. Milanfar, "Estimating Spatially Varying Defocus Blur From A Single Image", IEEE Trans. on image processing, volL. 22, no. 12, pp. 4879 - 4891, Dec. 2013.

[25] Y. Cao, S. Fang, and Z. Wang "Digital Multi-Focusing From a Single Photograph taken with an Uncalibrated Conventional Camera", IEEE Trans. on image processing, vol. 22, no. 9, Sept. 2013 Pages: 3703 - 3714.

[26] S.S.Praveen and P.R.Aparna,"Single Digital Image Multi-focusing to Point Blur Model Based Depth Estimation Using Point," Inter.Jour. of Eng.and Adv.Tech.(IJEAT), Vol.5 Iss.1, pp.77.81 ,Oct. 2015.

[27] A. Saxena, S. H. Chung, and A. Y. Ng. "Learning depth from single monocular images, " In Proc. Adv. Neural Inf. Process. Syst. 2005.